\documentclass[sigconf]{acmart}

\usepackage{graphicx}
\usepackage{multirow,tabularx}
\usepackage{epstopdf}
\usepackage{booktabs} 
\usepackage{textcomp}
\usepackage{xcolor,colortbl}
\definecolor{White}{rgb}{1,1,1}
\definecolor{Gray}{gray}{0.9}
\definecolor{LightCyan}{rgb}{0.88,1,1}

\usepackage{kantlipsum}     

\usepackage{hyperref}

\copyrightyear{2021}
\acmYear{2021}
\setcopyright{acmlicensed}\acmConference[GECCO '21 Companion]{2021 Genetic and Evolutionary Computation Conference Companion}{July 10--14, 2021}{Lille, France}
\acmBooktitle{2021 Genetic and Evolutionary Computation Conference Companion (GECCO '21 Companion), July 10--14, 2021, Lille, France}
\acmPrice{15.00}
\acmDOI{10.1145/3449726.3463197}
\acmISBN{978-1-4503-8351-6/21/07}

\begin{document}
\title{EBIC.JL - an Efficient Implementation of Evolutionary Biclustering Algorithm in Julia}

\author{Paweł Renc}
\affiliation{%
  \institution{AGH University of Science and Technology}
  \streetaddress{al. Mickiewicza 30 }
  \city{30-059 Krakow, Poland}
}
\email{rencpawe@gmail.com}

\author{Patryk Orzechowski}
\authornote{corresponding author}
\authornote{Patryk Orzechowski is also affiliated with AGH University of Science and Technology, Department of Automatics and Robotics, al. Mickiewicza 30, 30-059 Krakow, Poland}
\affiliation{%
  \institution{University of Pennsylvania}
  \streetaddress{3700 Hamilton Walk}
  \city{Philadelphia, PA 19104, USA}
}
\email{patryk.orzechowski@gmail.com}

\author{Aleksander Byrski}
\affiliation{%
  \institution{AGH University of Science and Technology}
  \streetaddress{al. Mickiewicza 30 }
  \city{30-059 Krakow, Poland}
}
\email{olekb@agh.edu.pl}

\author{Jarosław Wąs}
\affiliation{%
  \institution{AGH University of Science and Technology}
  \streetaddress{al. Mickiewicza 30 }
  \city{30-059 Krakow, Poland}
}
\email{jarek@agh.edu.pl}

\author{Jason H. Moore}
\affiliation{%
  \institution{University of Pennsylvania}
  \streetaddress{3700 Hamilton Walk}
  \city{Philadelphia, PA 19104}
}
\email{jhmoore@upenn.edu}

\renewcommand{\shortauthors}{P. Renc et al.}
\renewcommand{\shorttitle}{EBIC.jl - an Efficient Implementation of Biclustering Algorithm in Julia}

\begin{abstract}
Biclustering is a data mining technique which searches for  local patterns in numeric tabular data with main application in bioinformatics. This technique has shown promise in multiple areas, including development of biomarkers for cancer, disease subtype identification, or gene-drug interactions among others. In this paper we introduce EBIC.JL - an implementation of one of the most accurate biclustering algorithms in Julia, a modern highly parallelizable programming language for data science. We show that the new version maintains comparable accuracy to its predecessor EBIC while converging faster for the majority of the problems. We hope that this open source software in a high-level programming language will foster research in this promising field of bioinformatics and expedite development of new biclustering methods for big data.
\end{abstract}



\maketitle

\section{Introduction}

Evolutionary algorithms, since their introduction in the 1960s, have been primarily applied to global optimization tasks \cite{DBLP:journals/ml/GoldbergH88}. Later their applicability was extended to tackle multi-criteria optimization tasks \cite{DBLP:conf/gecco/ZitzlerD07},  where special care is needed to upkeep the diversity of the search space sampling in order to produce viable results. Evolutionary approaches have been also successfully applied to multiple machine learning tasks, including construction of decision trees \cite{ZHAO2007809}, feature selection \cite{4653480}, feature construction \cite{krawiec2002}, symbolic model identification \cite{doi:10.1080/10618600.2014.899236}, parameter estimation \cite{ZHOU2005451}, image feature manipulation \cite{10.1145/2425836.2425855}. In recent years, evolutionary algorithms have also gained much traction in the machine learning community, as multiple evolutionary approaches have led to competitive with state-of-the-art results in tasks of regression \cite{lacava2016epsilon,orzechowski2018symbolicreg}, deep learning models optimization \cite{8300639}, neural networks architecture selection   \cite{miikkulainen2019evolving, lu2019nsga} or automating machine learning (AutoML) \cite{Olson2016tpot}. A detailed survey on evolutionary machine learning may be found in \cite{doi:10.1080/03036758.2019.1609052}.

Another area of research in which evolutionary approaches have achieved leading results is biclustering. This unsupervised machine learning technique, also named co-clustering or subspace clustering, aims at discovering relevant local patterns in input data by extracting biclusters - submatrices with specific properties, e.g., with same values in certain rows or/and columns, correlated rows, or rows and columns, shift, scaled, or shift and scaled values in rows \cite{Madeira2004,Eren2013, Padilha2017, Orzechowski2018giga}. Since its first application to gene expression data over 20 years ago \cite{Cheng2000}, biclustering and its algorithms have led to some meaningful discoveries in biology and biomedicine, including identification of biomarkers for cancer \cite{tchagang2008early,wang2013biclustering}, diseases subtypes \cite{dao2010inferring,yang2017analysis,liu2014network} or adverse drug effects \cite{harpaz2011biclustering}.

The development of highly effective biclustering methods capable of capturing multiple patterns in a large volume of data is an emerging challenge. Up to this date tens or hundreds of biclustering methods have been proposed, including Plaid \cite{Lazzeroni2002}, ISA \cite{Bergmann2003}, Bimax \cite{Prelic2006}, cMonkey \cite{reiss2006cmonkey}, QUBIC \cite{Li2009}, FABIA \cite{Hochreiter2010}, Unibic \cite{wang2016unibic}, just to name a few. Not until 2018 has any of them been capable of providing high detection accuracy of multiple patterns, popular in biclustering research, in reasonable time. The landscape has changed with the introduction of Evolutionary search-based Biclustering (EBIC), which managed to solve all of the problems with average accuracy exceeding 90\%, and additionally offered unprecedented scalability \cite{Orzechowski2018ebic}. The algorithm, which takes its strength from GPU utilization, was originally dedicated for a single GPU but was later scaled to multi-GPU systems \cite{Orzechowski2018ebic2}. Implemented in highly performing C++ and CUDA, EBIC became a reference point for newly developed algorithms, such as QUBIC2 \cite{Xie2019qubic2} and RecBic \cite{Liu2020recbic}.

Limited availability of methods implemented in high productive programming languages is one of the main obstacles slowing down biclustering research. As the biclustering problem is considered NP-hard, the compromise between the ease of development and rapid prototyping vs high performance is hard to accomplish. This is the place where modern programming languages developed with a built-it parallelism paradigm come with help, as they allow fast and easy prototyping and testing improvements of new solutions, while not compromising on efficiency and running times.

Julia, an emerging programming language for Data Science developed at MIT, has been specifically designed for high-performance numerical computing.  It represents multi-paradigm language, uniting in itself features of different paradigms: functional, imperative and object-oriented. Julia programs compile using just-in-time (JIT) strategy to efficient native code using Low Level Virtual Machine (LLVM). Availability of multiple machine learning libraries, e.g., FLUX.jl~\cite{Flux.jl-2018}, and MLJ~\cite{Blaom2020}, as well as wrappers, e.g. TensorFlow.jl \cite{malmaud2018tensorflow}, ScikitLearn.jl \footnote{https://github.com/cstjean/ScikitLearn.jl} increase productivity in creating modern AI or machine learning applications. All aforementioned features make this specialized high-level programming language very convenient for rapid prototyping.

As Julia becomes more prevalent in the data science world, there is an emerging need for expanding the spectrum of available algorithms in this programming language. The release of multiple packages, such as CUDA.jl \cite{besard2018juliagpu, besard2019prototyping} allows to conveniently utilize NVidia graphics cards and further speed computation. After over 6 years since its first release in 2014, the library is mature and, provides, due to Julia's reflection and metaprogramming capabilities, an abstraction that allows direct interaction with CUDA API.

In this paper we present \emph{EBIC.jl}, an optimized and updated GPU-based implementation of EBIC biclustering algorithm in Julia. To our best knowledge, this might be the first available and thoroughly tested implementation of any biclustering algorithm in this modern programming language for data science. Our method is benchmarked on two large collections of datasets that have been previously used for measuring performance of biclustering methods. The results on 275 different datasets show that EBIC.jl offers higher adaptability to newly introduced sets of problems than the original algorithm (vanilla EBIC) and converges in multiple cases much faster and to better solutions. We also demonstrate that in vast majority of scenarios EBIC.jl is highly competitive with the leading method in the field, RecBic. 

\section{Methods}

In this section we present the design of the study and describe in more details two  collections of the datasets used for benchmarking, the methods included in this study as well as the metric, which was used to measure their performance.

\subsection{Datasets}

In order to deliver a more thorough picture describing performance of the newly developed method, two different benchmarks were used for evaluation of the performance: collection from Wang et. al  with 119 datasets grouped into 8 different sets of problems \cite{wang2016unibic}, which we called \emph{UniBic benchmark}, and collection from Xie et al. with 156 datasets grouped into 14 different sets of problems \cite{xie2018time}, which we named \emph{RecBic benchmark}. It was the authors' decision to reuse the existing datasets in the domain, instead of proposing another test suite, similarly to the authors of UniBic or RecBic.

\paragraph{Unibic benchmark} UniBic \cite{wang2016unibic} is a synthetically generated set of the most biologically meaningful types of data containing biclusters. It is divided into three test groups, each of which corresponds to a different problem commonly encountered in gene expression analysis.

The first one is evaluated to check an algorithm's ability to identify six popular data patterns: trend-preserving, column-constant, row-constant, shift, scale, and shift-scale, often referred  as Type I, Type II, Type III, Type IV, Type V and Type VI respectively. The tests assess how accurately algorithms detect differently-sized square biclusters in relatively small input data (up to $300 x 200$). The tests include three different sizes of biclusters, 5 replicates for each size, 90 datasets in total.

The second test set analyses the behavior of a method in scenarios with increasingly overlapping biclusters of size $25x25$. The patterns intersect up to $9x9$ elements creating four separate test cases with 5 replicates in each, 20 datasets in total.

The last set verifies biclustering methods' accuracy in very narrow pattern detection, i.e., ones with many rows and just several columns. The methods are requested to detect three biclusters with 10, 20, and 30 columns (3 test cases) and 100 rows implanted in the matrix of size $1000x100$.

The total number of datasets in this benchmark is 119.

\paragraph{RecBic benchmark}The second benchmark was downloaded from \cite{Liu2020recbic}. It consists of synthetically generated datasets with implanted biclusters, and similarly to the Unibic benchmark, the datasets are divided into several test groups. The tests for six different bicluster patterns are present in both benchmarks and were omitted by us in this one.

The benchmark comes with the following new challenges compared to the Unibic benchmark:
\begin{itemize}
\item \emph{Noise.} The test case validates algorithms' robustness by checking their ability to handle poor-quality  data. The resistance of the methods is tested under three different magnitudes of background noise (0.1, 0.2 and 0.3 with 0 being a sanity check).

\item \emph{Colincrease.} The test scenario verifies how well biclustering methods find a solution when biclusters the number of conditions constituting a biclusters increase (up to 120 columns).

\item \emph{1000Colin.} The test case examines how accurately algorithms discover biclusters when a background matrix is broadened from 500 to 2k columns, simultaneously maintaining the same size and position of true biclusters across all datasets.

\item \emph{Different.} In all other test cases, the data comes from Gaussian distribution with expectation 0 and deviation equal 1. In this test scenario, the influence of the various distributions of data constituting biclusters is checked by increasing the expectation, from 0 being a sanity check, up to 6.

\item \emph{20k-gene datasets.} Finally, to imitate real gene expression data in size, some of the aforementioned test cases (i.e., Six Types, Overlap, Different, Noise and Colincrease) were prepared in their larger variants with 20,000 rows and 250 columns.
\end{itemize}

Each test case comes with four variants resulting in the 156 datasets included, after removing cases overlapping with the previous benchmark.

\subsection{Performance measure}

Biclustering methods have been traditionally evaluated against ground truth using two performance measures: recovery and relevance \cite{Prelic2006}. Both measures are  based on the Jaccard index between the rows \cite{jaccard1912distribution} and reflect an average match of the maximum score for all rows of the biclusters to the ground truth. The relevance of biclusters shows how well detected biclusters represent true bicluster, whereas recovery shows the extent to which true biclusters are found by a biclustering method. 

A much more intuitive measure of biclustering performance which involves both rows and columns of biclusters was proposed by Patrikainen and Meila and is called Clustering Error (CE) \cite{patrikainen2006comparing}. The name of this metric is a bit confusing - it doesn't measure an error, but  the goodness of fit of two sets of biclusters. Thus, 1 indicates a perfect match and 0 denotes the worst possible fit. Before the metric is calculated, a confusion matrix is built based on the intersection of the biclusters, i.e. number of common elements shared between the detected and true  biclusters. An optimal assignment between two sets of biclusters is found using Munkres algorithm \cite{munkres1957algorithms}. Finally, the metric is calculated according to (\ref{eq:ce}):

\begin{equation}
\label{eq:ce}
CE(S,S_1)= \frac{|U|-D_{max}}{|U|}
\end{equation}

where $D_{max}$ is the maximal sum on diagonal of confusion matrix between the ground truth and  permutations of order of detected biclusters, and $U$ is the union of biclusters. 
One of the properties of CE is much heavier penalization compared to recovery, or relevance for incorrect assignments of biclusters' either rows, or columns. Clustering error was also demonstrated to have more desired properties to objectively measure performance of the methods in biomedical studies in comparison with other measures \cite{horta2014similarity,Padilha2017,Orzechowski2018giga}.

\subsection{Biclustering methods}

In this study the following methods served as a benchmark: EBIC, RecBic, QUBIC2 and Runibic (parallel implementation of Unibic).

\paragraph{EBIC}

EBIC \cite{Orzechowski2018ebic, Orzechowski2018ebic2} is an algorithm relying on evolutionary search which supports distribution of computations across multiple graphic cards (GPUs). It was inspired by the concept of detecting order-preserving patterns introduced in OPSM~\cite{BenDor2003}. Each individual in the population corresponds to a different bicluster, which is represented as a series of columns. The series defines an increasing order for the values in all of the rows that belong to the bicluster. The method uses simple genetic operations (such as insertion, deletion, substitution, swap, or crossover) in order to create new series of columns, or to merge two existing patterns. EBIC takes advantage from using GPUs in order to parallelize evaluation of fitness of the individuals. The evaluated scores are fetched in GPUs and transmitted back to CPU.

EBIC features multiple evolutionary strategies. Only a small fraction of best-fitting individuals is passed to the next generation (elitism), and its missing part is replenished by mutating other chromosomes. Tournament selection with exponential penalty from overusing same columns is used to increase diversity of the population (crowding). The individuals in new generations are prevented from being reevaluated by storing their hashes (tabu list). If a certain number of tabu list hits is recorded, the method finishes as it converged to the point where it is unwilling to search other solution subspaces. The enhancement prevents doing irrelevant iterations and hence remarkably reduces overall algorithm time. 

EBIC was demonstrated to be a highly accurate method, but also very scalable up to 8 GPUs thanks to its use of Nvidia CUDA API \cite{orzechowsk2019mining}. 

\paragraph{RecBic} RecBic \cite{Liu2020recbic} rephrases the concept of monotonously increasing trends of columns, which was successfully implemented in EBIC. RecBic starts with biclusters that contain 3 columns and exhaustively greedily expands their series by adding more columns, provided that there is sufficient number of matching rows. RecBic, similarly to EBIC, excludes rows that violate monotonicity and allows certain degree of noise. RecBic is based on not very scalable source code of QUBIC \cite{Li2009}.

\paragraph{QUBIC2} One of the recent biclustering methods dedicated to analyzing RNA-seq datasets is QUBIC2 \cite{Xie2019qubic2}. This modification of QUBIC biclustering algorithm originally developed by Li et al. \cite{Li2009} was demonstrated as the most versatile method across different platforms, including synthetic, microarrays, bulk RNA-Seq and scRNA-Seq datasets. QUBIC2, similarly to its predecessor QUBIC, uses graph representation of biclusters and tries to find heavy subgroups in a graph with vertices representing rows, and edge  weights - the level of similarity between the rows. The method comprises of three steps: discretization of input data, graph construction with seed selection, building core biclusters, which are further expanded and filtered. The source code of QUBIC2 is also based on its predecessor QUBIC.

\paragraph{Runibic} Runibic~\cite{Orzechowski2018runibic}, another method that conceptually attempts to identify monotonous trends in the dataset, is a parallelized version of UniBic biclustering method by Wang et al. \cite{wang2016unibic}. The method relies on evaluation of the longest common subsequence (LCS) of ranks between multiple pairs of rows. The ranks are determined by the increasing order of the values in each row, where the rank of $k$ reflects the $k$-th smallest element in each row. Although the source code of its predecessor is based on QUBIC, Runibic fixes some of its limitations by providing new and more modern implementation in R with C++ backend.

\subsection{EBIC.jl}

EBIC.jl proposed herein is a novel implementation of EBIC in Julia, and it is probably the first highly effective biclustering method available in this programming language. The selection of technology remarkably expedited the development process and enabled to locate issues due to the significantly lessened code complexity compared to the previous version. The algorithm works as follows: 

\begin{enumerate}
    \item Initialize population with randomly generated short chromosomes represented by a vector of column numbers.
    \item Score every individual of the initial population by compressing the whole population and sending it on GPU where trends between rows are sought.
    \item Download numbers of trend-preserving rows corresponding to each chromosome, evaluate scores and update the top rank list.
    \item Initialize a new population with best-fitting individuals from the top rank list (elitism).
    \item Replenish the new population with mutated chromosomes from the old population penalizing for frequently used columns.
    \item Evaluate tabu list hits (the number of mutations after which the resulting chromosome was rejected) and if the count reaches a threshold, go to step 9.
    \item Score population and update the top rank list.
    \item Replace the old population with the newly created one and return to step 4.
    \item Send all individuals from the top rank list on GPU and download corresponding trend-preserving rows' indices. The pairs of sets of rows and columns create biclusters.
\end{enumerate}

Additionally, to increase the performance, the following modifications were incorporated compared to the original EBIC:

\paragraph{Using registers and atomic operations}
Most of the optimizations that might have influenced the performance are strictly related to CUDA kernels. Firstly, shared memory used for storing local variables was replaced by registers as their access time is much shorter \cite{orzechowski2015effective}. A significant change was not making an intermediate data storing point when acquiring GPU computations results and exploiting atomic operations, which are thread-safe.

\paragraph{Loop unrolling} Subsequently, the reduce design pattern was improved by applying the loop unroll. This technique is a simple and popular method of optimization in GPU computing. It enables to omit loop control overheads such as end-of-loop tests and branch penalties by expanding a loop to regular code with control statements. The improvement could not be applied entirely because, in the current version of CUDA.jl, there is no correspondence of the volatile keyword from CUDA C. This is an indication for a compiler that the access to the shared memory must be an actual memory read or write instruction. With its use, there would not be a need for additional expensive memory synchronization.

\paragraph{Other differences to 'vanilla' EBIC} Finally, the values of constant parameters were further tuned. The allowed number of tabu hits was reduced as the method, even though converging quickly, was looking further for solutions in irrelevant subspaces. Additionally, we adjusted the initial population settings, as a significant part of the randomly generated chromosomes was spawned unreasonably short and died in the first cycle of evolution. It quickly resulted in a poorly differentiated generation delaying its proper development.

The new prototype EBIC.jl does not support multiple GPUs at this point. Although this limitation still allows to analyze pretty large sizes of datasets, it is a notable disadvantage for big data analyses, limiting the size of the datasets to available memory of a single GPU. The feature certainly will be the subject of further works.

The input parameters of the method include options of switching on/off negative trends, setting a threshold for approximate trends, the level of overlap allowed between the trends, number of iterations and biclusters. All the parameter of the method are documented on the Github repository.

\section{Results}

The selected biclustering methods were tested on the Unibic and RecBic benchmarks and their results quality was assessed using the Clustering Error (CE). Each of the test cases in these benchmark contains five replicates (datasets with the same characteristics). Therefore, due to a large number of test sets, initiated the method with multiple random seed was not deemed necessary. The biclustering experiments were conducted on two different workstations. CPU-based methods  (i.e., RecBic, QUBIC2 and Runibic) were tested on a machine equipped with Intel Core i7-10700 CPU\footnote{https://www.cpubenchmark.net/cpu.php?cpu=Intel+Core+i7-10700+\%40+2.90GHz\&id=3747}, whereas tests of GPU-based methods (i.e., EBIC and EBIC.jl) were carried on the workstation equipped with a bit slower Intel Core i7-9700KF CPU\footnote{https://www.cpubenchmark.net/cpu.php?cpu=Intel+Core+i7-9700KF+\%40+3.60GHz\&id=3428} and GeForce GTX 1050 Ti GPU. In our opinion, this experiment settings enables us to assess the methods' actual performance more objectively, as the former group benefits more from CPU multithreading, whereas the latter group from GPU parallelization.

The settings of the benchmarked algorithms are presented in Table \ref{table:settings}. The methods were run with the settings suggested by the authors. For Runibic we tried both settings (default and legacy) and adopted one yielding better results on average (i.e. legacy). For QUBIC2, parameter N was used, which according to the algorithm's authors, refers to 1.0 biclustering (1.0 objective function + regular expansion) and in our trial runs yielded the best results on average. Additionally, all algorithms were requested to return exact number of biclusters, as implanted, for each test case.

\begin{table}[ht!]
\centering
\caption{Settings of the methods.}
\label{table:settings}
\begin{tabular}{c c } 
 \hline
 Algorithm & Settings \\ [0.5ex] 
 \hline
 EBIC.jl & -n 20000  \\ 
 EBIC & -n 20000  \\ 
 RecBic & [default] \\ 
 Runibic & [dafeault] and useLegacy=True  \\ 
 QUBIC2 & -N \\ 
 \hline
\end{tabular}
\end{table}

\subsection{Unibic benchmark}

\begin{figure}[hb!]
\begin{center}
\includegraphics[width=0.49\textwidth]{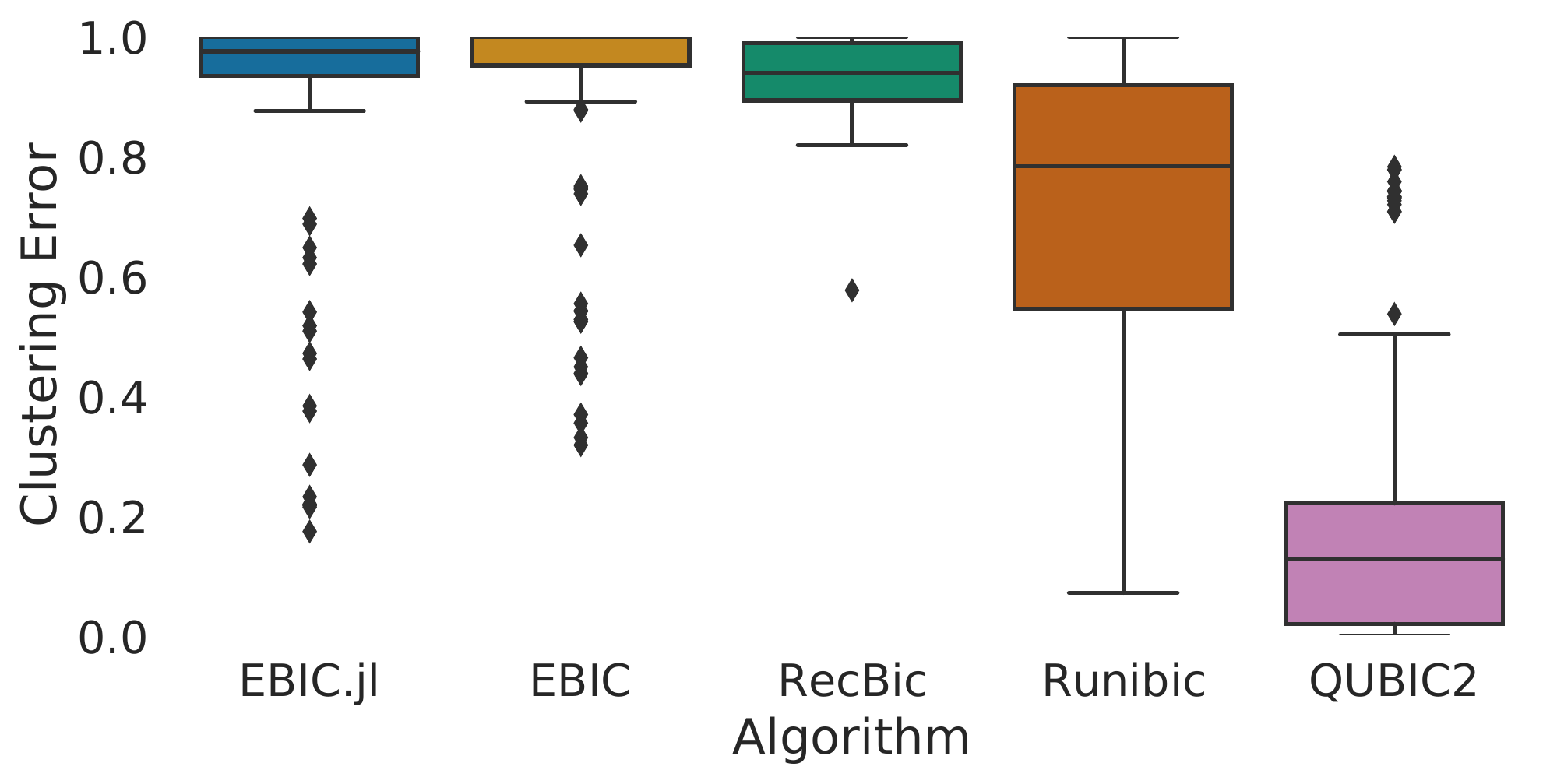}
\end{center}
\caption[Speedup]{Clustering Error by an algorithm measured on Unibic benchmark.}
\label{fig:ce_by_alg_unibic}
\end{figure}

Performance of the methods across multiple problems in UniBic benchmark in terms of CE is presented in Figure~\ref{fig:ce_by_alg_unibic}. EBIC achieves the best score overall for most tested datasets with its median equal $1$, whereas EBIC.jl and RecBic perform slightly worse, reaching $0.98$ and $0.94$ respectively. The median accuracy of Runibic in terms of CE on UniBic dataset is $0.78$. QUBIC2 gives remarkably lower results ($0.13$). 

\begin{figure*}[h]
\begin{center}
\includegraphics[width=\textwidth]{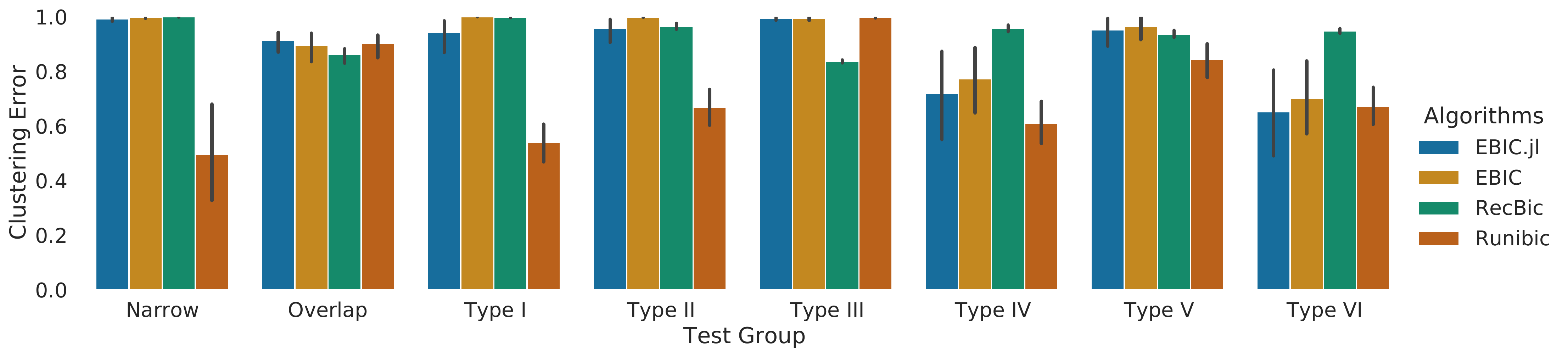}
\end{center}
\caption[Speedup]{Clustering Error by different test groups for Unibic benchmark. The higher score, the better.}
\label{fig:ce_by_test_unibic}
\end{figure*}

It is worth noting that both EBIC and EBIC.jl have difficulties delivering satisfactory results in specific test cases, which is visible in Figure~\ref{fig:ce_by_test_unibic}. The patterns of Type IV and VI, namely shift-scale and scale, are the most challenging for these algorithms, resulting in mean CE $0.77$ and $0.70$ for EBIC and $0.71$ and $0.65$ for EBIC.jl respectively. In the same test cases, RecBic discovers high-quality biclusters, scoring $0.95$ for both patterns, but instead, its performance is lower in the Overlap and Type III (row-const pattern) tests, reaching $0.86$ and $0.84$, which are worse than for EBIC ($0.89$, $0.99$), EBIC.jl ($0.91$, $0.99$) and Runibic ($0.90$, $0.99$).

Additionally, Runibic's average CE score is $0.74$, and QUBIC2, apart from the Overlap test case, cannot find any accurate solution in the considered benchmark.

\subsection{RecBic benchmark}

The methods were subsequently tested on the second collection of the datasets. It needs to be noticed that this collection of datasets is provided by the authors of one of the methods included in the comparison (RecBic). Thus, some bias resulting in higher than expected performance of RecBic might be present in the following analysis.

\begin{figure}
\begin{center}
\includegraphics[width=0.49\textwidth]{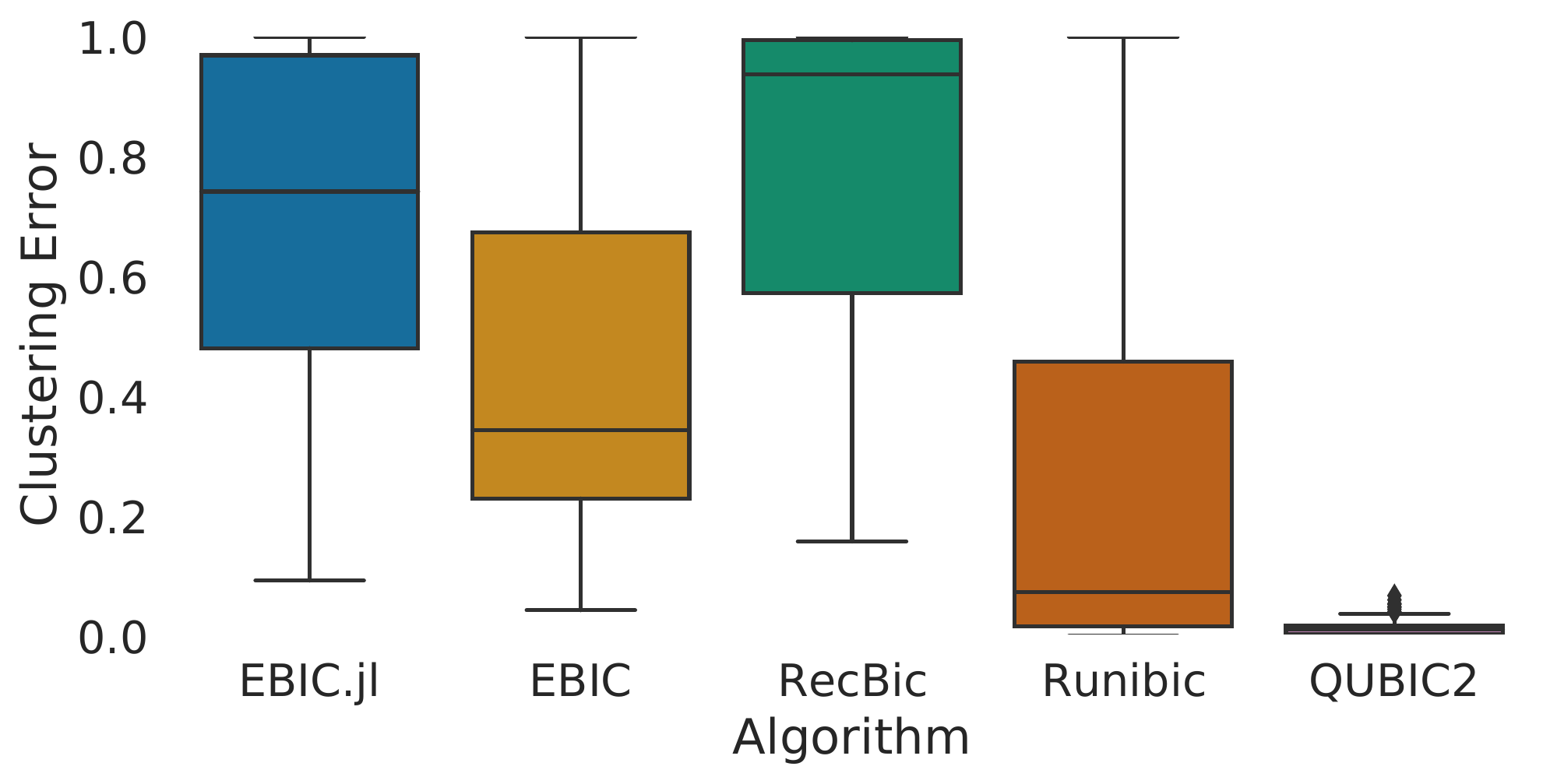}
\end{center}
\caption[Speedup]{Clustering Error by an algorithm measured on RecBic benchmark. The higher, the better.}
\label{fig:ce_by_alg_recbic}
\end{figure}

As shown in Figure~\ref{fig:ce_by_alg_recbic}, RecBic achieves the best performance, standing out from the competing methods with a median equal to $0.94$. The second best algorithm is EBIC.jl with a median quality of $0.74$. The remaining methods visibly underperform, with medians $0.34$ (EBIC), $0.07$ (Runibic) and $0.01$ (QUBIC2). It should be noted that the small performance difference on the previously evaluated benchmark between EBIC.jl and its ancestor (EBIC) no longer holds, confirming that the newly incorporated improvements brought desired outcomes.

\begin{figure*}
\begin{center}
\includegraphics[width=\textwidth]{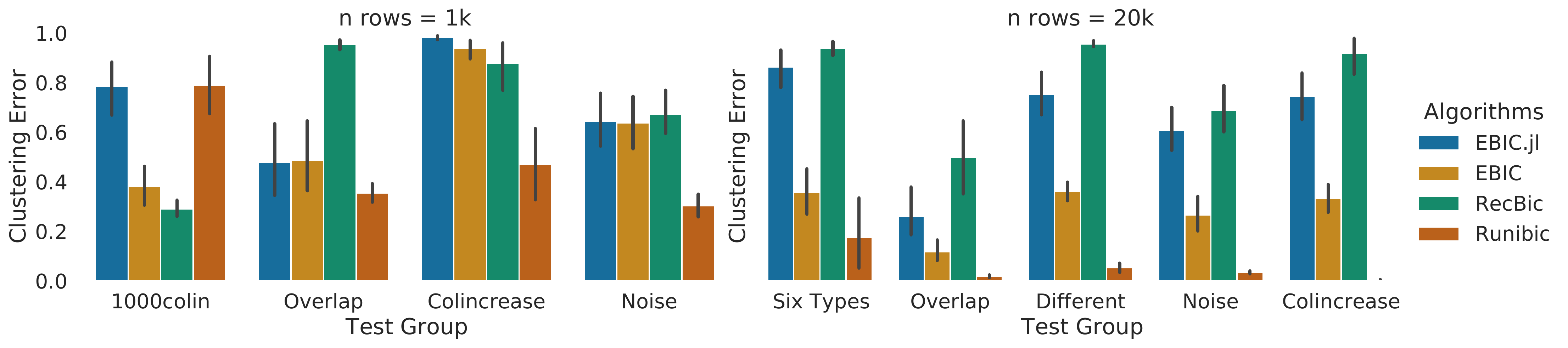}
\end{center}
\caption[Speedup]{Clustering Error for different methods by different test groups on RecBic benchmark. To the left: test cases with 1,000 rows, to the right: test cases with 20,000 rows. The higher score, the better.}
\label{fig:ce_by_test_recbic}
\end{figure*}

For the smaller datasets with 1,000 rows, RecBic outperforms the rest of the algorithms in all tests (Fig.~\ref{fig:ce_by_test_recbic}) except  \emph{Colincrease} and \emph{1000colin} scenarios, where EBIC.jl performs better. EBIC.jl seems to be less affected by the increasing number of columns than RecBic. Large number of columns improves the performance of Runibic in both scenarios. EBIC.jl remains also the only method that can stand up to RecBic for the datasets with 20,000 columns, although its performance is lower.

Two \emph{Overlap} scenarios need additional comment. The design of this scenario and placement of the biclusters seems to clearly put RecBic in favor. During the inspection of the heatmap of the datasets, we have unintentionally pointed out the same biclusters, as were found using EBIC and EBIC.jl, which turned out to be incorrect ones, both in number and location. We believe it is debatable which biclusters should be yielded in this scenario, but we also understand the rationale of the solutions found by RecBic, which in this scenario are closer to the ground truth.

Across the benchmark, Runibic obtains mediocre results for the majority of 1k-gene datasets at CE level equal $0.46$ on average. Runibic runs into issues for 20k-gene datasets, as it is unable to find satisfactory solutions and crashes in two test cases of \emph{Colincrease}, not outputting any results. QUBIC2 does not find any sensible solution in all of the RecBic benchmark datasets and achieves a  mean score of CE of  $0.01$.

Afterwards analysing the overall results of the benchmarked we dwelved more thoroughly into specific test groups.

\begin{figure}
\begin{center}
\includegraphics[width=0.49\textwidth]{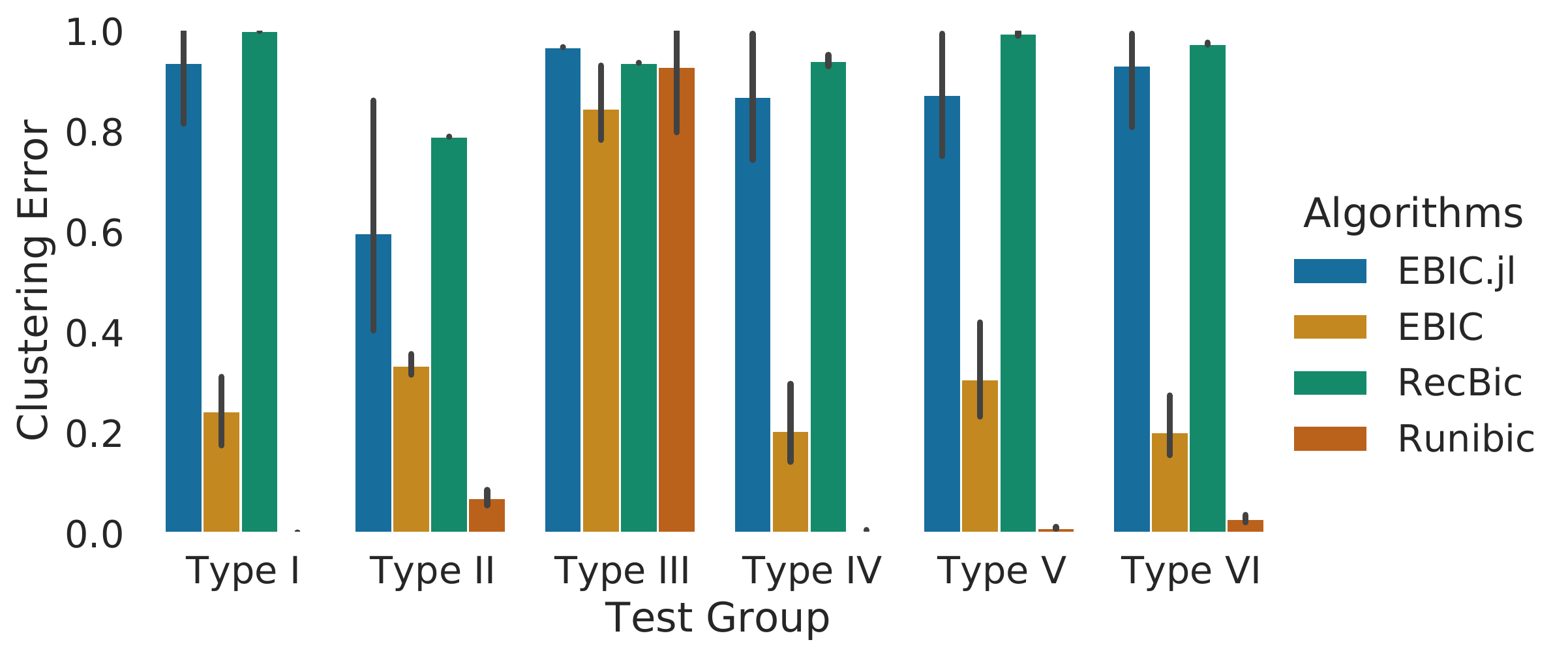}
\end{center}
\caption[Speedup]{Clustering Error of the algorithm for six biclustering patterns on RecBic benchmark. Notice high increase in performance on this benchmark of EBIC.jl vs original EBIC. The higher score, the better.}
\label{fig:ce_6types_recbic}
\end{figure}

The results of \emph{Six Types} are shown in Figure~\ref{fig:ce_6types_recbic} broken down into distinct patterns. The test scenario is a much larger version of the six patterns from the Unibic benchmark. Type III scenario, namely row-constant pattern, turns out to be the easiest dataset for all the considered biclustering methods except for QUBIC2. For this scenario, EBIC.jl achieves the average CE of $0.97$, which is slightly better than the other methods: RecBic $0.94$, Runibic $0.93$ and EBIC $0.84$. For all the remaining patterns, RecBic turns out to discover the best-quality biclusters, achieving $0.94$ on average on all scenarios. This score is higher than the second best method, EBIC.jl which average CE score is equal to $0.84$. The remaining methods struggle to perform well on the benchmark, EBIC -- $0.26$, Runibic -- $0.02$ and QUBIC2 -- $0.02$. It can also be observed that the margin between EBIC.jl and RecBic on Type IV and VI patterns (shift-scale and scale) is much lower than in Unibic benchmark. 

\begin{figure}
\begin{center}
\includegraphics[width=0.49\textwidth]{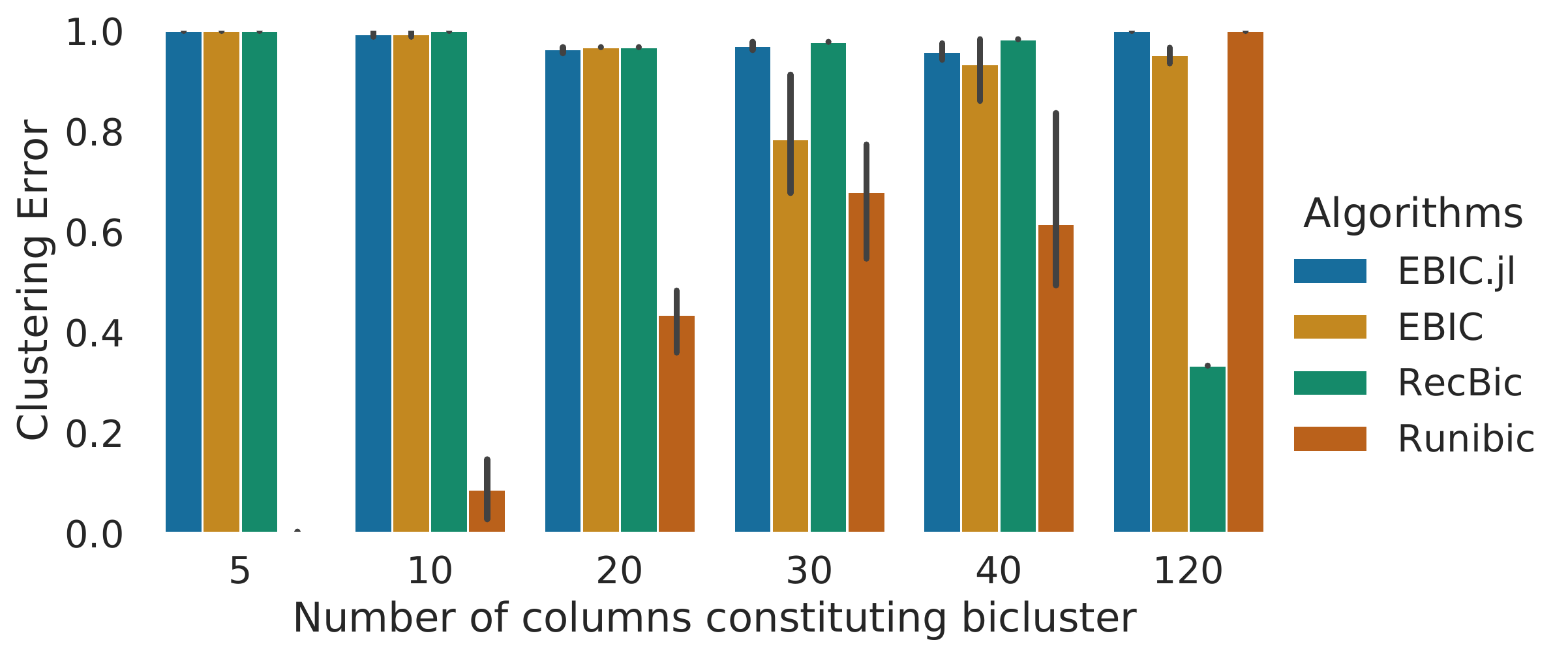}
\end{center}
\caption[Speedup]{Clustering Error of biclustering with increasing number of columns on RecBic benchmark (\emph{Colincrease} scenario). Notice that the performance of RecBic visibly drops with 120 columns, whereas EBIC.jl and Runibic maintain high accuracy.}
\label{fig:ce_colincrease_recbic}
\end{figure}

Subsequently, we break down the outcomes of the \emph{Colincrease} test group in its 1k-gene variant (see Figure~\ref{fig:ce_colincrease_recbic}). EBIC.jl, EBIC and RecBic discover almost perfectly narrow and medium-wide biclusters reaching nearly $1.0$ CE in most cases, only EBIC's accuracy slight decreases for some datasets in the 30-column test. However, in the wide-biclusters scenario, RecBic is incapable of finding an acceptable solution achieving merely $0.33$ CE. On the contrary, EBIC.jl maintains high accuracy across different widths of the biclusters,  scoring CE of $0.98$ in all test cases on average. Runibic is utterly ineffective in discovering narrow biclusters, but its performance improves with the increasing number of columns reaching $1.0$ CE in the broad-biclusters test case. It confirms that this method was purposefully designed  for such scenarios. QUBIC2 again does not come with any acceptable solution.

\begin{figure}
\begin{center}
\includegraphics[width=0.47\textwidth]{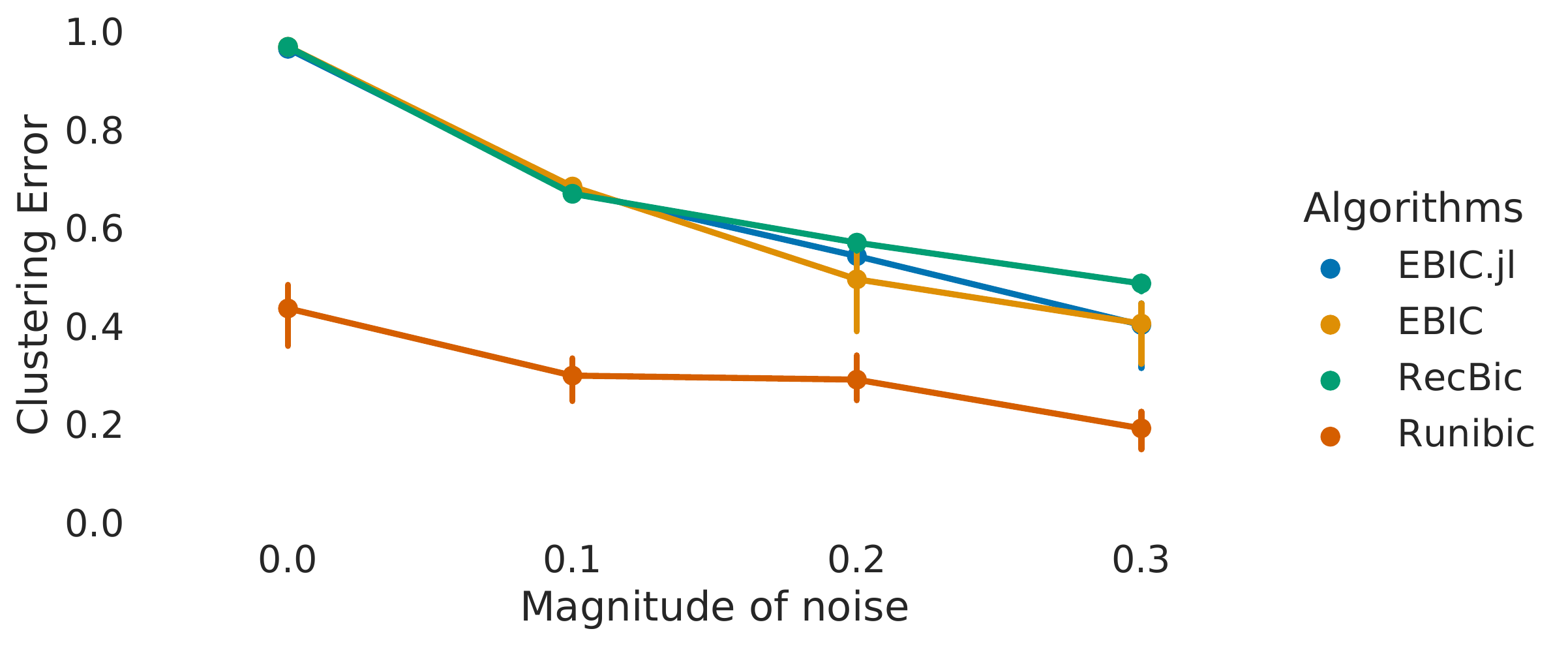}
\end{center}
\caption[Speedup]{Clustering Error of biclustering methods for noise scenario on RecBic benchmark.}
\label{fig:ce_noise_recbic}
\end{figure}

Finally, we analyze how different noise levels incorporated in genes' data interfere with the performance of the biclustering methods. The results can be observed in Figure~\ref{fig:ce_noise_recbic}. The first test case involves a sanity check with no noise included, the best three algorithms (EBIC.jl, EBIC and RecBic) achieve the same CE equal to $0.97$. Then, together with increasing noise, their efficiency almost identically deteriorates, reaching CE $0.48$ for RecBic, and $0.4$ for EBIC.jl and EBIC in the scenario of the highest tested noise. It demonstrates that RecBic is somewhat more resistant when dealing with poor-quality data. Runibic seems to react similarly to the algorithms mentioned above, but its performance is much lower than the leading methods ($0.3$ CE on average). Lastly, we can not evaluate how noise in data influences QUBIC2 as the quality of all its results is close to $0.01$ CE on average.

\subsection{Running times}

It should be stressed that the capability of finding high-quality solutions is essential, but a method applied in real-life scenarios is also requested to provide results in reasonable amount of time. The choice of the most suitable method might impose a compromise by trading overall accuracy for notably shorter execution time. For that reason, we compared the execution times of all five considered biclustering methods to check how fast they are in various scenarios. For this purpose, we again used two benchmarks, Unibic and RecBic.

Unibic/Runibic was historically deemed the leading algorithms in the field of biclustering. Similarly, QUBIC2 was claimed to outperform multiple other methods (including EBIC and runibic) on diverse collection of datasets \cite{Xie2019qubic2}. However, as shown in the previous Subsection, on two large collections of datasets, they were found to be incapable of detecting high-grade solutions in the majority of test scenarios. In order to better analyze differences in performance of the current leading methods (RecBic, EBIC and EBIC.jl) we have focused on those methods. 

\begin{figure}
\begin{center}
\includegraphics[width=0.47\textwidth]{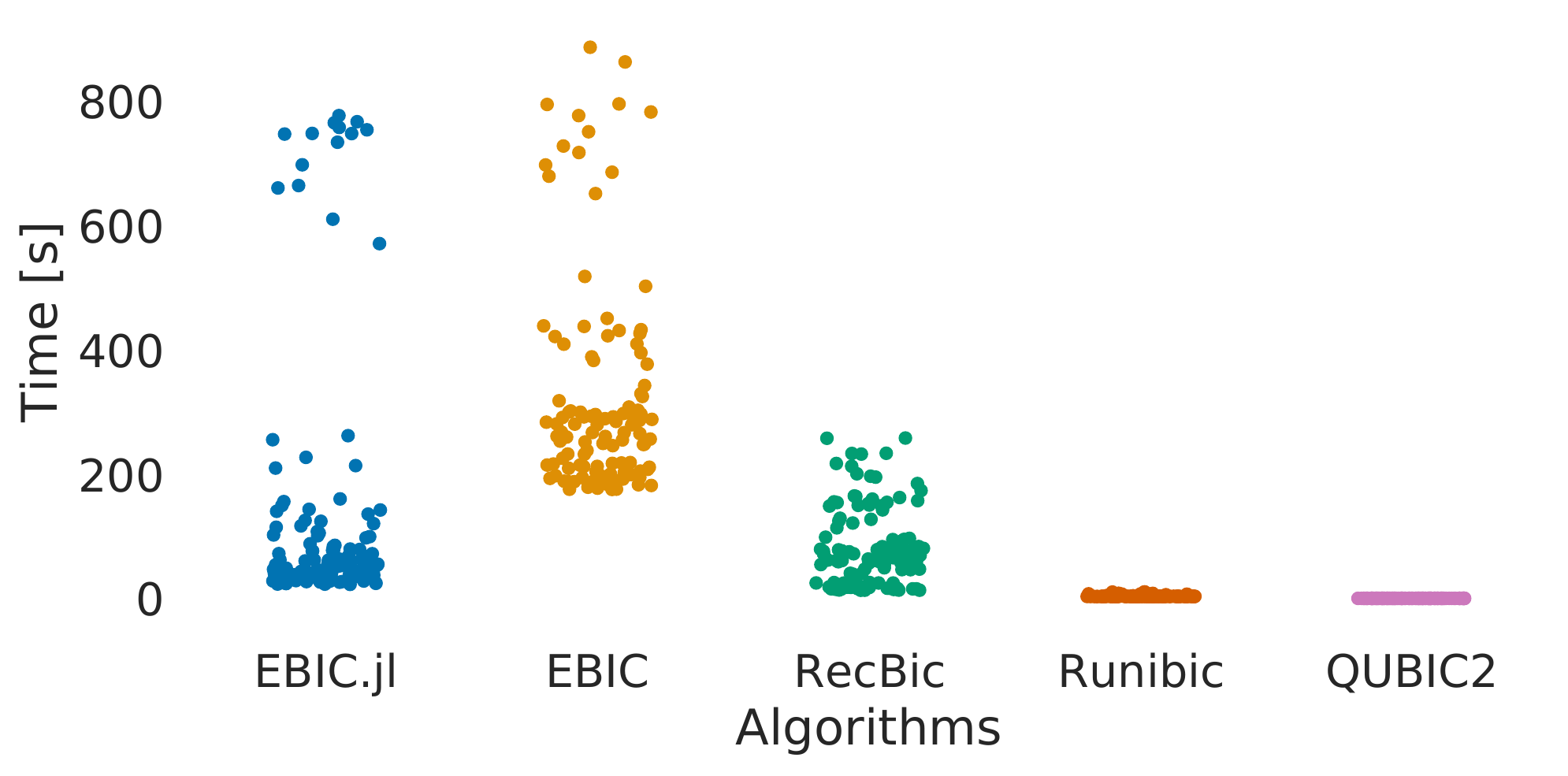}
\end{center}
\caption[Speedup]{Running times by algorithm on Unibic benchmark. Each dot represents a single dataset.}
\label{fig:time_by_alg_unibic}
\end{figure}

\begin{figure}
\begin{center}
\includegraphics[width=0.47\textwidth]{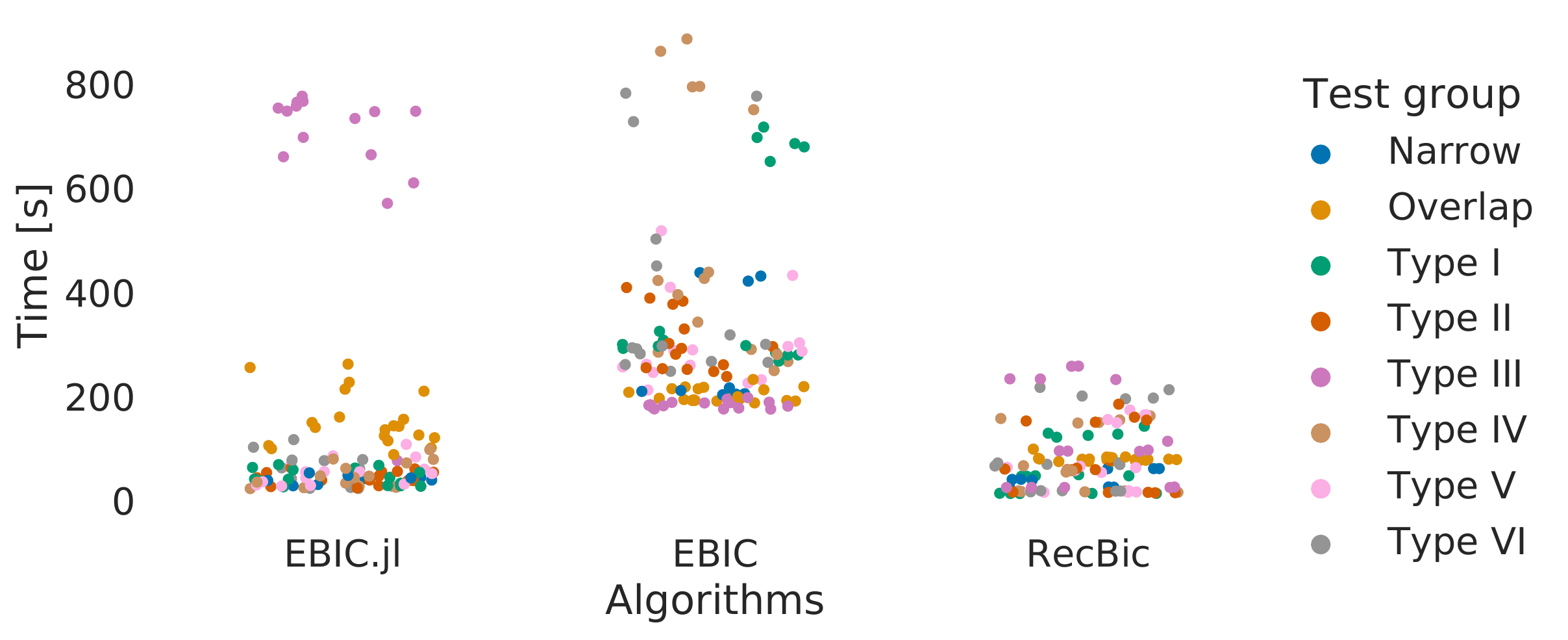}
\end{center}
\caption[Speedup]{Running times by algorithm on Unibic benchmark for three top biclustering methods with the test group division. Each dot represents a single dataset.}
\label{fig:time_by_alg_unibic_top}
\end{figure}

The execution times obtained on the Unibic benchmark are presented in Figures~\ref{fig:time_by_alg_unibic} and \ref{fig:time_by_alg_unibic_top}. Its datasets are relatively small, ranging from 150 to 1000 rows and a few tens of columns. Among the more precise methods, RecBic operates the quickest in all test cases, lasting only 81 seconds on average. Our algorithm usually tackles most of the problems somewhat longer, apart from a few cases. The test cases of Type III, namely row-const pattern, take surprisingly long for EBIC.jl (11 minutes on average), pushing its average execution time up to 145 seconds. On the other hand, EBIC, even though being a predecessor for our algorithm, achieves worse performance (5.5 minutes on average) and operates visibly slower in different test cases than EBIC.jl (e.g. Type IV -- 8.3 minutes and Type I -- 7 minutes on average). Runibic and QUBIC2 are both very fast algorithms, their average time is 3.6 and 0.02 seconds respectively.

\begin{figure}
\begin{center}
\includegraphics[width=0.47\textwidth]{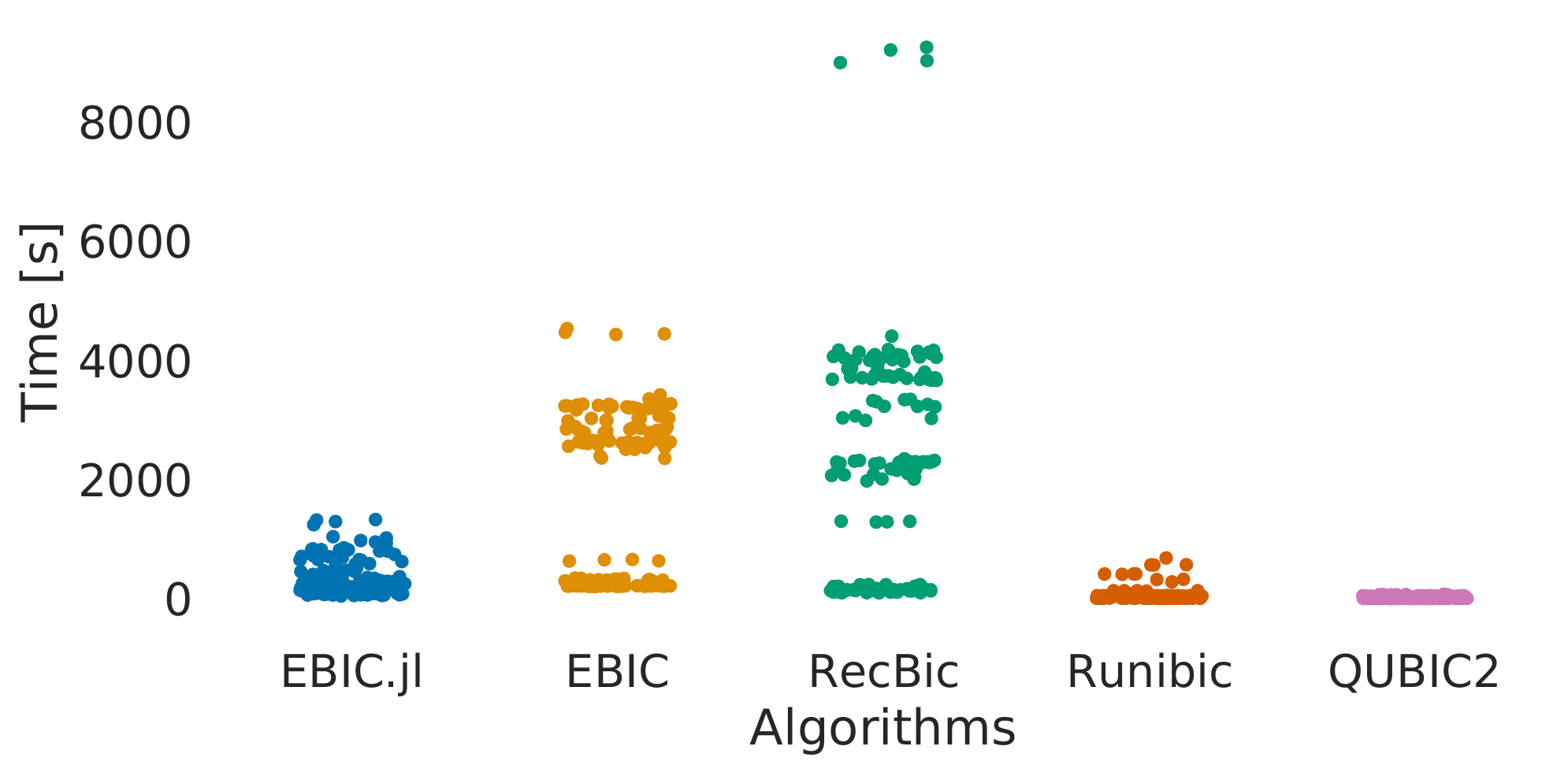}
\end{center}
\caption[Speedup]{Running times of algorithms on RecBic benchmark. Each dot represents a single dataset.}
\label{fig:time_by_alg_recbic}
\end{figure}

\begin{figure}
\begin{center}
\includegraphics[width=0.47\textwidth]{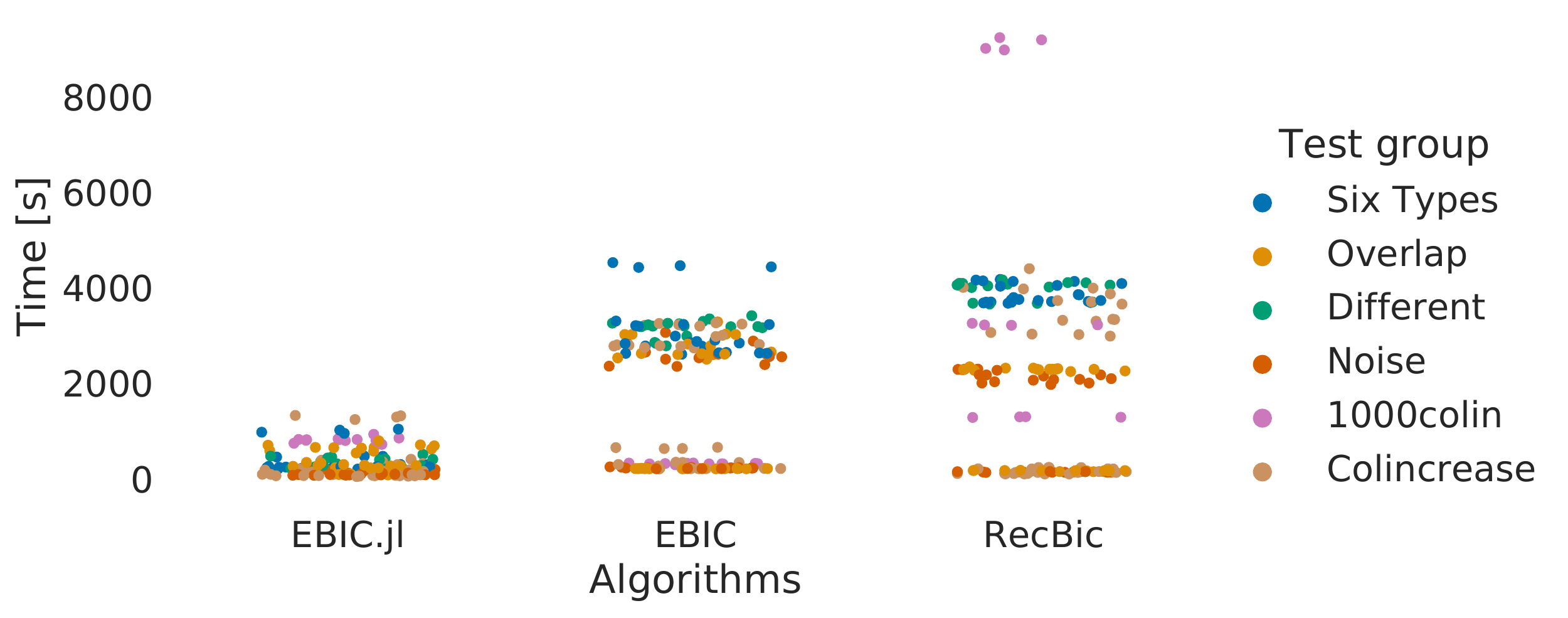}
\end{center}
\caption[Speedup]{Running times of algorithms on RecBic benchmark for three top biclustering methods with the test group division. Each dot represents a single dataset.}
\label{fig:time_by_alg_recbic_top}
\end{figure}

The second comparison is made on RecBic benchmark, which includes two sizes of datasets (1k rows and 20k rows). The execution time of all runs is presented in Figures~\ref{fig:time_by_alg_recbic} and \ref{fig:time_by_alg_recbic_top}, showing the running times across all of the problems.

Among the most precise biclustering algorithms, EBIC.jl discovers biclusters in the shortest time on average (248s), which is considerably shorter than two other algorithms, EBIC (1148s) and RecBic (1294s). The most time-consuming test case for the RecBic algorithm is \emph{1000Colin}, with 2k columns because the algorithm, even though operating 2.5 hours on average, scores only $0.3$ CE, whereas our algorithm's average time for this scenario is 13 minutes, and it achieves $0.59$ CE. This comparison suggests that our method is much more scalable to larger datasets than RecBic, and greatly lower running times are somewhat compromised by slightly lower accuracy. Additionally, the less precise algorithms again run much faster then these mention above. Their average times in the benchmark are: 1 minute for Runibic and 21 seconds for QUBIC2. 

\section{Conclusions}
This paper introduces \emph{EBIC.jl}, an open-source implementation of GPU-based evolutionary biclustering algorithm EBIC in Julia. To our best knowledge, this is the very first available and working biclustering method in this emerging programming language for data science. The proposed method was benchmarked on two extensive collections of datasets: introduced by Wange et al. \cite{wang2016unibic} with 119 datasets grouped in eight different sets of problems, and Liu et al. \cite{Liu2020recbic} with 156 datasets and nine problems, and achieved the accuracy competitive to the state-of-the-art methods.

Our second contribution is providing a modern, optimized and enhanced implementation of one of the leading biclustering methods in the field, which not only works faster than the original method, but also outperforms it across multiple scenarios -- visible especially in the second benchmark. There are several reasons, why EBIC.jl in Julia works faster than the original EBIC implemented in C++. First,  the amount of communication sent between GPU and CPU was reduced by changing organization of CUDA kernel. Thus, EBIC.jl sends  to CPU  a single value for each of the columns in the group of threads. Reorganization also allowed us to use loop unrolling, which is considered more efficient. Second, the tabu hits allowance was decreased, what resulted in faster convergence. Third, we improved the generation of initial population, which was not efficiently managed in the original EBIC. Some other issues were also fixed, e.g. the genetic mutations did not work as expected and there was no chance for one for the deletion to happen. 

This study has certain limitations. First, we have decided not to evaluate the methods on gene expression datasets. The reason for that is that the final results very frequently depend on how the datasets were preprocessed. As none of the workflows has been established in biclustering, our study might unnecessarily bring high bias and become detached from the previous reports. Secondly, statistical analyses comparing performance of the methods were excluded from the paper. The main reason for that is that such analyses are not commonly applied in biclustering benchmarking studies (e.g. Eren et al. \cite{Eren2013}, Padilha et al. \cite{Padilha2017}). Instead, in line with the previous studies, a detailed report highlighting performance of the methods on solving different biclustering problems is provided. Thirdly, the comparison of the runtimes between EBIC and EBIC.jl might be biased, as some optimizations (loop unroll, atomics etc.) have not been behchmarked for the vanilla EBIC. 

Overall, we believe EBIC.jl can be considered a worthy expansion, or even a successor of EBIC. The new implementation has around half less lines of code than the original one and marks a milestone for prototyping, proving that an application written in a high-level programming language can successfully compete in performance with those written in considered the most efficient language, C++. Simultaneously ultimately excelling them in terms of software engineering, debugging, maintenance and potential extensibility. 

Future work on the method will focus on expanding the implementation to multiple GPUs and parallelization of CPU parts of the code. 
The method will also be tested on multiple public datasets.
Lastly, we plan to work on incorporating new techniques that will accelerate the algorithm's converging time.

The new implementation will also serve as a base for development of new biclustering methods. With already fast and parallel GPU-supported evaluation of the results, the emphasis can now be placed on generation of new candidates which can potentially lead to achieving better final solution in faster time.

\section*{Algorithm availability}
EBIC.jl is open source: \url{https://github.com/EpistasisLab/EBIC.jl}

\begin{acks}
This research was supported in part by PLGrid Infrastructure and by NIH grants LM010098 and LM012601.
\end{acks}

\section*{Authors' contribution}
PO conceived the study. PR implemented the algorithm and performed analyses. PR and PO analyzed the results. PR and PO wrote the draft of the manuscript with help of AB, JW and JHM. AB and JW consulted the project. All the authors edited and reviewed the manuscript. JHM supervised the project.

\bibliographystyle{ACM-Reference-Format}
\bibliography{sample-bibliography}

\onecolumn

\end{document}